\documentclass{article}

\usepackage{arxiv}

\usepackage[utf8]{inputenc} 
\usepackage[T1]{fontenc}    
\usepackage{hyperref}       
\usepackage{url}            
\usepackage{booktabs}       
\usepackage{array}
\usepackage{amsmath}
\usepackage{amssymb}
\usepackage{amsfonts}       
\usepackage{nicefrac}       
\usepackage[final,nopatch=footnote]{microtype}       
\usepackage{lipsum}
\usepackage{graphicx}
\usepackage{cite}
\usepackage{xcolor}         
\usepackage{dsfont}
\usepackage{pifont}
\usepackage{tabularx}
\usepackage{multirow}
\usepackage{threeparttable}
\usepackage{caption}
\usepackage{subcaption}
\usepackage{makecell}
\usepackage{dblfnote}
\usepackage{xspace}

\newcommand{\ie}{\textit{i.e.}\xspace}
\newcommand{\eg}{\textit{e.g.}\xspace}
\newcommand{\etal}{\textit{et al.}\xspace}

\title{Mitigating the Impact of Prominent Position Shift in Drone-based RGBT Object Detection}

\author{
 Yan Zhang \\
  School of Computer Science\\
  Wuhan University\\
  Wuhan {\rm 430072}, China \\
  \texttt{zhangyan@whu.edu.cn} \\
   \And
 Wen Yang \\
  School of Electronic Information\\
  Wuhan University\\
  Wuhan {\rm 430072}, China \\
  \texttt{yangwen@whu.edu.cn} \\
  \And
 Chang Xu \\
  School of Architecture, Civil and Environmental Engineering\\
  École Polytechnique Fédérale de Lausanne\\
  Lausanne {\rm 1015}, Switzerland \\
  \texttt{xuchangeis@whu.edu.cn} \\
  \And
 Qian Hu \\
  School of Electronic Information\\
  Wuhan University\\
  Wuhan {\rm 430072}, China \\
  \texttt{huq1an@whu.edu.cn} \\
  \And
 Fang Xu \\
  School of Computer Science\\
  Wuhan University\\
  Wuhan {\rm 430072}, China \\
  \texttt{xufang@whu.edu.cn} \\
  \And
 Gui-Song Xia \\
  School of Computer Science\\
  Wuhan University\\
  Wuhan {\rm 430072}, China \\
  \texttt{guisong.xia@whu.edu.cn} \\
}

\begin{document}
\maketitle
\begin{abstract}
Drone-based RGBT object detection plays a crucial role in many around-the-clock applications. However, real-world drone-viewed RGBT data suffers from the prominent position shift problem, \ie, the position of a tiny object differs greatly in different modalities. For instance, a slight deviation of a tiny object in the thermal modality will induce it to drift from the main body of itself in the RGB modality. 
Considering RGBT data are usually labeled on one modality (reference), this will cause the unlabeled modality (sensed) to lack accurate supervision signals and prevent the detector from learning a good representation. Moreover, the mismatch of the corresponding feature point between the modalities will make the fused features confusing for the detection head.
In this paper, we propose to cast the cross-modality box shift issue as the label noise problem and address it on the fly via a novel Mean Teacher-based Cross-modality Box Correction head ensemble (CBC).
In this way, the network can learn more informative representations for both modalities. 
Furthermore, to alleviate the feature map mismatch problem in RGBT fusion, we devise a Shifted Window-Based Cascaded Alignment (SWCA) module. SWCA mines long-range dependencies between the spatially unaligned features inside shifted windows and cascaded aligns the sensed features with the reference ones. 
Extensive experiments on two drone-based RGBT object detection datasets demonstrate that the correction results are both visually and quantitatively favorable, thereby improving the detection performance.
In particular, our CBC module boosts the precision of the sensed modality ground truth by 25.52 aSim points. Overall, the proposed detector achieves an $\mathrm{mAP}_{50}$ of 43.55 points on RGBTDronePerson and surpasses a state-of-the-art method by 8.6 $\mathrm{mAP}_{50}$ on a shift subset of DroneVehicle dataset.
The code and data will be made publicly available.
\end{abstract}


\section{Introduction}
RGBT object detection is crucial in many around-the-clock applications such as security and search \& rescue. Most existing methods assume that image pairs are geometrically aligned. However, even image pairs registered by alignment algorithms are weakly aligned in practice~\cite{zhang2019weakly}, which leads to position shifts between the same object in RGB and thermal modalities. Especially in drone-viewed scenarios, prominent position shifts pose substantial challenges to multimodal representation learning of tiny objects.

\begin{figure*}[htpb]
\centering
  \includegraphics[width=0.79\linewidth]{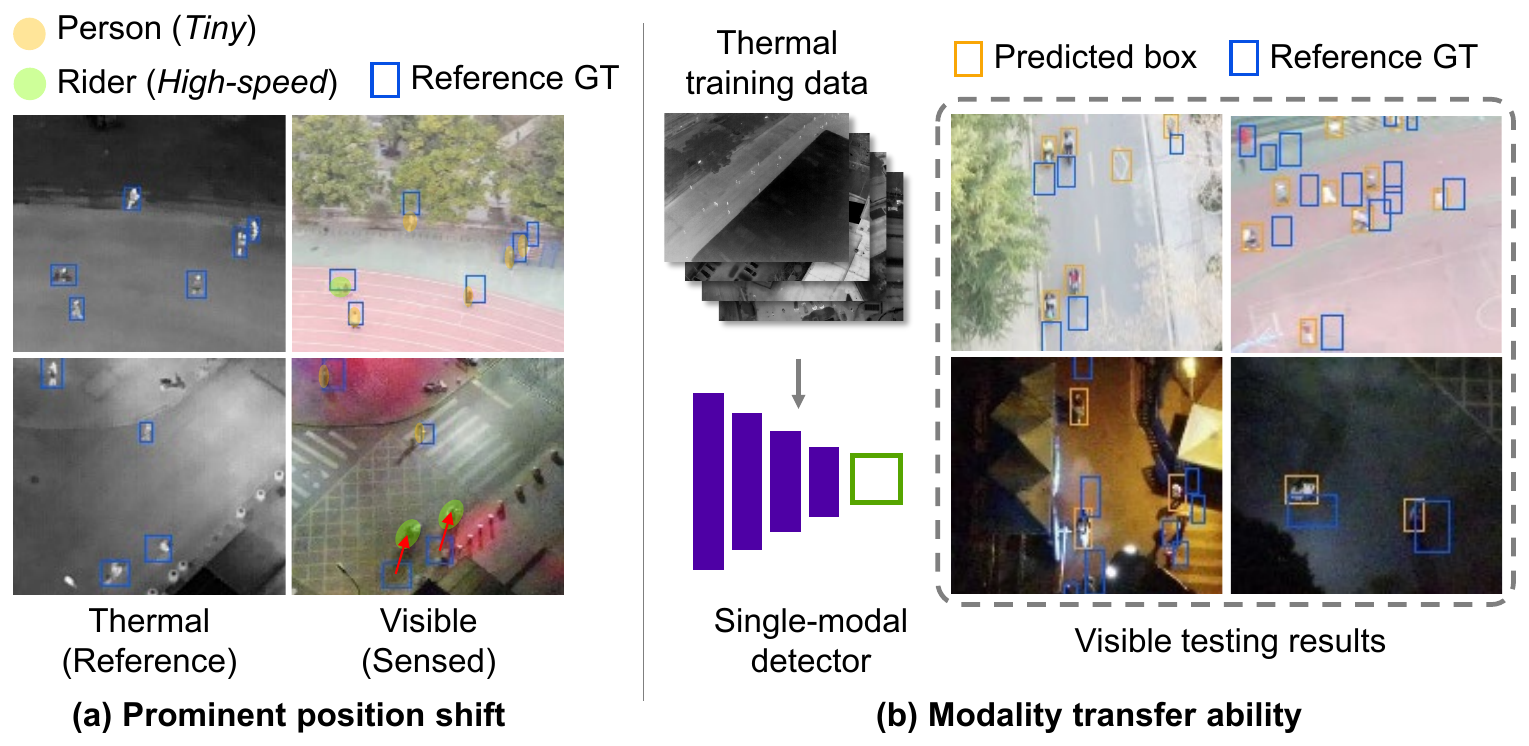}
  \caption{(a) Demonstration on prominent position shifts. ``Reference'' denotes the modality with aligned GTs; ``Sensed'' denotes the one without GTs.  Yellow and green masks indicate the main body of the person and rider objects. (b) The detection results on the visible modality of a thermal-trained detector. This well represents the modality transferability between the RGB and thermal modalities.
  }
  \label{fig:fig1}
\end{figure*}

The position shift problem originates from (1) \textbf{discrepant physical properties of different sensors}: The positions on hardware, field-of-views, and resolutions of RGB and thermal sensors are discrepant. Although there are some works~\cite{kaist,jia2021llvip} designing special hardware for calibration, specially customized imaging sensors are not applicable in many real-world scenarios because of the additional cost and low availability; (2) \textbf{time asynchronization}: especially when capturing fast-moving objects, due to the inevitable imaging time lag, the moving objects will have large position shifts in different modalities. The position shifts may also originate from the camera movement. When the cameras are moving, due to time asynchronization, the same object will present a position shift in different modality images. 
Moreover, in drone-based RGBT imagery, a slight deviation of a tiny object in one modality will make it drift away from the main body of itself in the other modality, making the position shift (3) \textbf{prominent for tiny objects}~\cite{zhang2023drone}. Fig.~\ref{fig:fig1} (a) illustrates the phenomenon of prominent position shifts. The position shifts of riders are especially large as riders move at a high speed (See red arrows in the bottom right image). 

The prominent position shift is an unignorable issue in drone-based RGBT object detection, especially for tiny objects. First of all, it will cause (1) \textbf{bounding box confusion}. The annotated bounding box drifts from the object in the visible modality, making the detector struggle to learn object representation in the visible modality. Further, it could lead to (2) \textbf{severe feature map mismatch}. The features of different modalities are misaligned in corresponding positions, which could make the fused features confusing to the detection head. These issues will heavily degrade the performance of the RGBT object detectors.

The position shift problem was first analyzed by Zhang \etal~\cite{zhang2019weakly}, where the concept of reference and detected images is first introduced into the multispectral setting. Besides, they propose AR-CNN, which fixes the reference modality and performs alignment on the sensed one. This work is further extended to oriented object detection in remote sensing imagery by~\cite{yuan2022translation}. 
Napat \etal~\cite{napat2021misalignment} point out that the previous works are constrained in weak misalignment and propose a multi-modal pedestrian detector for large misalignment. These existing works alleviate the position shift problems to some degree. However, most of them require accurate and paired annotations for both modalities to learn the shift pattern, which is labor-intensive. Although Napat \etal~\cite{napat2021misalignment} study the large position shift problem, it is limited on manually generated modality shifts instead of real-world data. Manually generated shifts are fixed globally, while real-world position shifts vary in direction and distance even within one image pair, as shown in Fig.~\ref{fig:fig1} (a).
Different from previous works, we focus on the prominent position shift, especially for tiny objects, where the RGBT features of objects are easily aliased. Moreover, we use the ground truth (GT) of one modality to derive those of the other modality, providing auxiliary supervision without additional labeling costs. 

Since the RGBT data are typically labeled on one modality, we define the labeled modality as the \textit{reference modality} and the unlabeled modality as the \textit{sensed modality}. This definition takes the perspective of ``labeled \& unlabeled'', instead of simply ``fixed \& unfixed''~\cite{zhang2019weakly}. The annotated GT boxes are aligned with the reference modality, while in the sensed modality, these GT boxes are not spatially accurate but somewhat indicate the coarse positions of objects. Another observation is that there is abundant shared low-level (edge) and high-level (semantic) information between the visible and thermal modalities. Based on the shared features, a detector trained with thermal modality (reference) can yield proper responses on visible modality (sensed) without domain adaptation techniques (See Fig.~\ref{fig:fig1} (b)). Considering the inherent discrepancy between the modalities, we can introduce domain adaptation techniques to alleviate this problem and better exploit shared information.

Motivated by the aforementioned observations, we propose to view the shifted GT boxes in the sensed modality as inaccurate boxes and manage to correct them on the fly. In this paper, we design a novel framework to adaptively correct the shifted boxes for the sensed modality and align multi-modal features considering long-range cross-modality dependencies. Firstly, we devise a Cross-modality Bbox Correction (CBC) module to correct shifted GT boxes in the sensed modality. We take inspiration from the Mean Teacher~\cite{tarvainen2017mean} framework in the unsupervised domain adaptation (UDA)~\cite{kennerley20232pcnet} setting to leverage knowledge from the reference modality, thus improving the correction performance. As shown in Fig.~\ref{fig:overall} up, our framework employs a teacher head to correct shifted boxes for the sensed modality (sensed GT), and a student head takes reference and sensed modalities as inputs and is supervised by sensed GTs and reference GTs, respectively. We take reference GTs as initialization for sensed GTs. The weights of the teacher's head are updated by the Exponential Moving Average (EMA) of student weights. Therefore, the correction results of the teacher are improved as the student is trained. In terms of box correction, we design a multiple instance learning (MIL)~\cite{liu2022robust,wu2023spatial} inspired GT box correction method. For each inaccurate GT box in the sensed modality, we select its positive samples together with their confidence scores to construct an instance bag. Then combined with the high-quality instances, we derive the corrected GT box from the previous one. Corrected GT boxes are updated epoch by epoch to achieve better representation learning for the sensed modality. In particular, CBC is applied only at the training stage and does not affect inference speed. 

CBC alleviates the bounding box confusion problem while the feature map mismatch still exists. Previous works~\cite{zhou2020improving} rely on deformable convolution to adaptively align the multi-modal features. However, under prominent position shifts, the receptive field of a convolution layer is limited to capturing semantic relations between modalities. Moreover, the size of small objects in aerial view makes the alignment more challenging. Imprecise deformation of features could easily spoil the object features, thereby bringing down the detection accuracy. In this work, we propose Shifted Window-based Cascaded Alignment (SWCA) to align sensed features with reference features. 
A SWCA block predicts offsets based on window multi-head cross-attention between sensed and reference embeddings. Then the sensed features are deformed based on predicted offsets. We cascade two SWCA blocks and introduce shifted window design~\cite{liu2021swin}. After SWCA, we fuse the aligned RGBT feature for the subsequent detection head.

Our contribution can be summarized as follows.
\begin{itemize}
    \item We analyze the cause and influence of the prominent position shift problem in drone-based RGBT object detection. Further, we propose diminishing the bounding box confusion from the perspective of the correction of the bounding box. Our CBC module achieves adaptive alignment without any additional annotations and automatically produces better annotations for the sensed modality on the fly. With better GT signals for the sensed modality, the detector can learn more informative representation and enhance the multi-modal detection performance.
    \item We alleviate the influence of the feature map mismatch by SWCA. Taking advantage of the cross-attention mechanism, we manage to build semantic connections between mismatched RGBT feature maps and predict better offsets for sensed modality. By devising a shifted window-based cascaded alignment scheme, we achieve delicate feature deformation without spoiling the features of tiny objects.
    \item We conduct experiments on two drone-based RGBT object detection datasets, RGBTDronePerson~\cite{zhang2023drone} and DroneVehicle~\cite{sun2022drone}. Consistent improvements on the two challenging and distinct datasets show that our method is robust to real-world prominent position shifts and effective in utilizing multi-modal information. In addition,  we propose a metric named aSim to evaluate the position shift of objects. The correction performance is shown to be both visually and quantitatively favorable.
\end{itemize}

\section{Related Work}
\label{sec:related}
\subsection{RGBT Object Detection}
Object detection is a fundamental task in computer vision. However, general object detection based on RGB images is vulnerable under unsatisfactory illumination conditions. The introduction of a thermal modality provides rich information around the clock and greatly improves the ability against low-light conditions. Therefore, extensive research has been conducted on RGBT object detection. 

\textbf{Fusion-based methods.} Some~\cite{iafrcnn,guan2019fusion,zhang2023tinet} propose to utilize the illumination condition to determine a reliable modality. Some researchers~\cite{kim2021uncertainty,sun2022drone,li2023multiscale} introduce uncertainty or confidence to fuse RGBT features accordingly. Others~\cite{zhang2019cross,zhou2020improving,shen2024icafusion,yuan2024c} resort to designing attention-based fusion networks to achieve adaptive fusion. Progress has been made in multi-modal feature fusion to combine the strengths of different modalities.

\textbf{Alignment-based methods.} 
However, these methods all assume that the multispectral image pair is geometrically aligned, which is impractical in the real world. 
\textit{(1) Weakly misalignment.} Zhang \etal~\cite{zhang2019weakly} first study the impact of the position shift problem in RGBT object detection and propose AR-CNN to align RoI region features. TSRA~\cite{yuan2022translation,yuan2024improving} extends AR-CNN to oriented object detection, taking into account angle and scale deviations. These methods manage to predict the shift pattern for each instance but demand paired multi-modal annotations, which is labor-intensive. C$^2$Former~\cite{yuan2024c} utilizes the pairwise correlation modeling capability of the Transformer to adaptively obtain calibrated and complementary features. OAFA~\cite{chen2024weakly} predicts the offset of features based on the common subspace learning of RGBT data. The above two methods do not require paired annotations, but rely on implicit calibration. 
\textit{(2) Large misalignment.} Napat \etal~\cite{napat2021misalignment} point out that the previous works are limited in weak misalignment and propose a modal-wise regression and a multi-modal IoU to tackle the large misalignment situation. However, they study large misalignments only under manually generated position shifts instead of real-world RGBT shifts, which consist of different directions and distances even in one image pair. 
\textit{(3) Prominent position shift.} This problem is first proposed by drone-based RGBT tiny person detection~\cite{zhang2023drone}. Different from vehicle-viewed RGBT misalignment, where most objects remain overlapped in different modalities, for example, a 5-pixel position shift will cause the reference GT box to drift away from the sensed tiny objects in drone view. QFDet~\cite{zhang2023drone} simply relies on the pooling layer to alleviate the impact of position shift on the RGBT features, which is insufficient.
In our work, CBC-Head only requires annotations for the reference modality to perform RGBT detection and automatically yields better annotations for the sensed modality. With explicit supervision, our method manages to learn informative RGBT representations.

\subsection{Learning with Noisy Annotations}
To improve the ability of deep neural networks against noise in annotations has been studied in various computer vision tasks. In image classification, various methods~\cite{yi2019probabilistic,zheng2021meta} are developed to identify noisy labels and further correct them. In object detection, the impact of label noise is more severe and complex since this task performs label classification and bounding box regression simultaneously. Chadwick \etal~\cite{chadwick2019training} firstly study the impact of noisy labels in object detection and propose a co-teaching method to mitigate the effects of label noise in object detection. Li \etal~\cite{li2020learning} propose a cleanliness score to mitigate the influence brought by noisy anchors. Liu \etal~\cite{liu2022robust} introduce multiple instance learning (MIL) to address the inaccurate bounding boxes problem. Wu \etal~\cite{wu2023spatial} also utilize the MIL technique but propose a spatial self-distillation based object detector to exploit spatial and category information simultaneously. 

These methods effectively improve the robustness of detectors under real-world noisy environments and simulated noisy settings. Nevertheless, a fundamental assumption of these methods is that there are many reliable and high-quality annotations to supervise the network. In our condition, we view the shifted bounding boxes as noisy annotations for the sensed modality. In this way, all of the annotations are noisy, making the problem distinct and challenging. 

\subsection{Unsupervised Domain Adaptation}
Unsupervised domain adaptation~\cite{li2022cross,he2022cross,kennerley20232pcnet,chen2023confidence,zhai2024maximizing} aims to learn transferable features to reduce the discrepancy between a labeled source and an unlabeled target domain. The task is more challenging in object detection because of the discrimination between objects and backgrounds. Mean Teacher~\cite{tarvainen2017mean} is proposed for semi-supervised learning. AT~\cite{li2022cross}, TDD~\cite{he2022cross}, and 2PCNet~\cite{kennerley20232pcnet} introduce the Mean Teacher framework into the UDA setting to improve the performance of cross-domain detection. In our problem setting, the reference modality is labeled while the sensed modality is unlabeled. Therefore, we take the reference modality as the source domain and the sensed as the target domain.

\section{Methodology}
\label{sec:meth}

\begin{figure*}[t]
  \centering
  \includegraphics[width=1.0\linewidth]{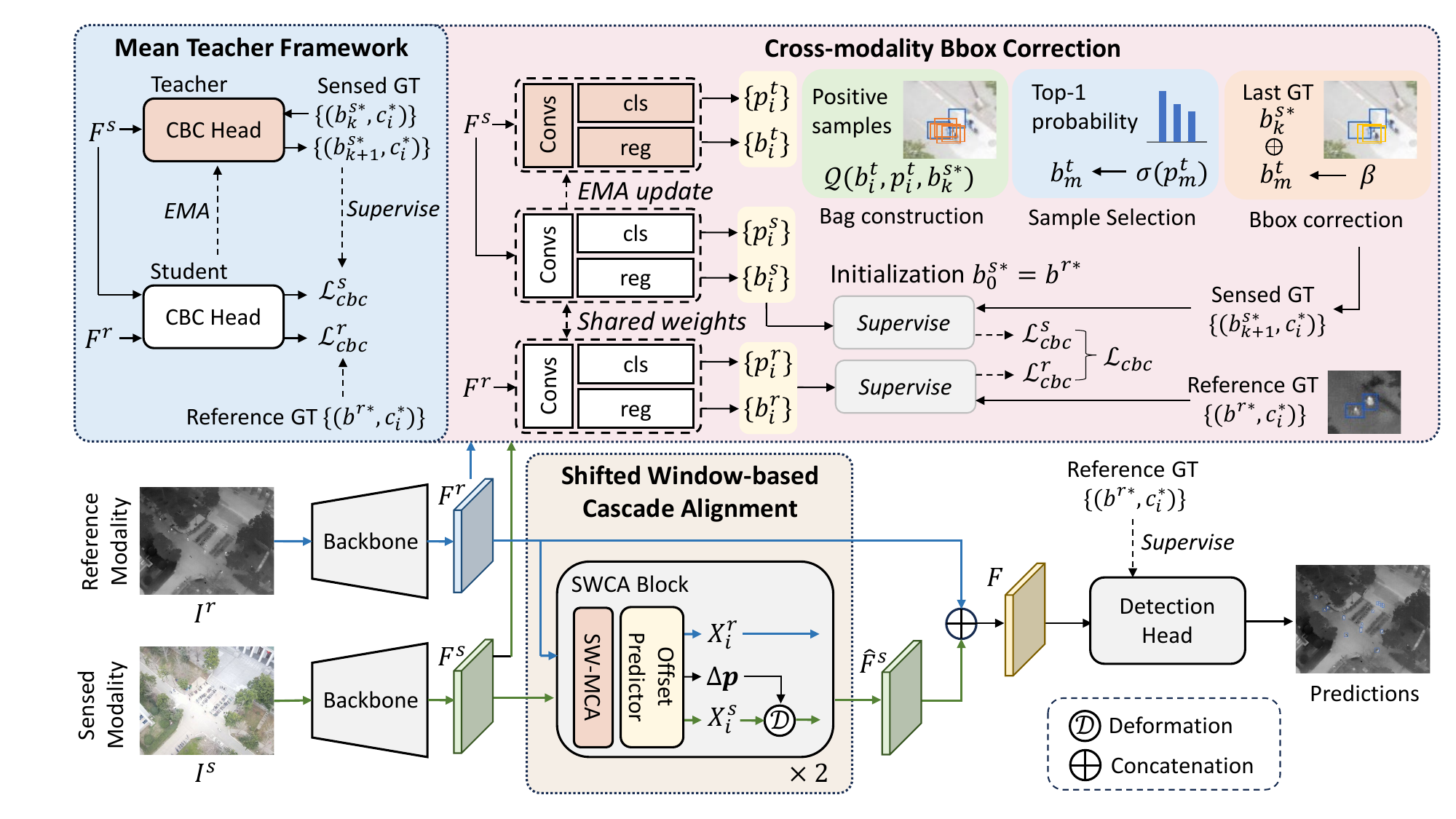}
  \caption{The overall structure of the proposed method in the training stage. The mainstream of the scheme is an RGBT fusion-and-detection scheme (down). Considering the prominent position shift in the sensed modality, we design a Cross-modality Bbox Correction (CBC) module under a Mean Teacher framework (up). The teacher CBC head takes the sensed feature $F^s$ as input and yield sensed GTs $\{(b^{s*}_{k+1},c^*_{k+1})\}$ via a cross-modality bbox correction strategy. The strategy consists of three steps, namely bag construction, sample selection, and bbox correction. Finally, the student CBC head is updated by the supervised reference loss $\mathcal{L}^r_{cbc}$ and the ``unsupervised'' sensed loss $\mathcal{L}^s_{cbc}$. The teacher weight is updated by the exponential moving average (EMA) of student weight.
  }
  \label{fig:overall}
\end{figure*}

Our work aims to obtain precise objects and informative representations in RGBT object detection under prominent position shifts. We define the modality with aligned annotations as the reference modality and the modality with prominent position shifts as the sensed modality. Given the two modalities and one set of training annotations, we try to yield better-aligned annotations for the sensed modality and finally obtain informative fused multi-modal features for detection. We will first introduce the overall structure of our framework in Sec.~\ref{sec:overall}. Then we will dive into the details of Cross-modality Bbox Correction and Shifted Window-based Cascaded Alignment, respectively, in Sec.~\ref{sec:cbc} and Sec.~\ref{sec:swca}.

\subsection{Overall Framework}
\label{sec:overall}
Fig.~\ref{fig:overall} illustrates the overall framework of our method in the training stage. We design a Cross-modality Bbox Correction module to produce GT boxes with better accuracy for the sensed modality and further enhance the learned representation of the sensed modality. The CBC module is constructed under a Mean Teacher framework and is only adopted in the training stage. Furthermore, we devise a Shifted Window-based Cascaded Alignment module to align the sensed feature with the reference feature. Finally, detections are made on the fused features.

\subsection{Cross-modality Bbox Correction}
\label{sec:cbc}

Under the circumstance of prominent position shifts, the detector struggles to learn meaningful representation with inaccurate GT supervision provided for the sensed modality. Furthermore, previous studies have shown that the multi-modal model tends to bias towards the dominated modality with better performance \cite{peng2022balanced}. These two challenges coupled together make the representation learning for the sensed modality more difficult. Without well-learned representation, it is hard to fully exploit the potential of RGBT data. Therefore, we propose to correct the GT boxes of the sensed modality so that the sensed representation will be more informative.
Since the thermal and visible modalities share abundant low-level and high-level information, we assume a modality adaptive detection head can yield reasonable responses to objects whether in the thermal modality or in the visible modality. 

\textbf{Mean Teacher framework.} We adopt an additional detection head ensemble to perform the cross-modality bounding box correction. The head ensemble is constructed under a Mean Teacher~\cite{tarvainen2017mean} framework similar to the UDA~\cite{kennerley20232pcnet} setting. Specifically in our problem setting, the reference modality is similar to the source domain, and the sensed modality is the target domain. The reason for introducing the UDA technique is that it brings in the knowledge from the reference modality and facilitates accurate predictions on the sensed modality so that the corrected GTs are more reliable. The head ensemble consists of two heads, teacher and student, which are identical in structure. The teacher head takes the sensed features $F^s$ as input and yields updated sensed GTs $\{(b^{s*}_{k+1},c^*_{k+1})\}$ via our cross-modality bbox correction strategy.
The student CBC Head is a shared head for the reference and sensed modality, where a common paradigm of classification and regression is adopted: features $F^m$ are fed to the classification branch and the regression branch, yielding classification and regression predictions $\{p_i^m\}$ and $\{b_i^m\}$. Given corresponding targets of bounding box coordinates $\mathcal{B}^{m*}=\{b^{m*}_i\}$ and their category labels $\{c^{*}_i\}$, the classification and regression losses are calculated by:
\begin{equation}
\begin{aligned}
&\mathcal{L}_{cls}^m=\frac{1}{N_{cls}}\sum_i^{N_{cls}}L_{cls}(p_i^m, c^*_i), \quad
&\mathcal{L}_{reg}^m=\frac{1}{N_{reg}}\sum_i^{N_{reg}}c_i^*L_{reg}(b_i^m, b^{m*}_i),
\end{aligned}
\end{equation}
where $N_{cls}$ and $N_{reg}$ denote sample numbers of classification and regression, respectively; $L_{cls}$ and $L_{reg}$ denote specific loss functions, \eg, cross-entropy loss for classification and L1 loss for regression; $m\in\{r,s\}$ denotes the modality, where $r$ for reference and $s$ for sensed. 

Since GT boxes $\{b^{r*}_1, b^{r*}_2, …, b^{r*}_n\}$ are aligned for the reference modality, with the above training loss supervision, the head gains a reliable ability to detect objects in a reference image. Furthermore, given that its paired sensed image shares much information in terms of edges and semantics, the predictions made on the sensed image have considerable value. Therefore, we propose to exploit the positive samples to correct the sensed GT boxes $\mathcal{B}^{s*}_k$ (epoch $k$) iteratively by epochs. With reference to \cite{wu2023spatial}, we introduce three steps for our sensed GT boxes correction: (1) bag construction: for each GT box $b^{s*}$, we collect its positive samples $(\{b^{s}_i,\mathrm{score}_i\})$ and form a proposal bag; 
(2) sample selection: for each proposal bag we select samples with scores higher than an adaptive threshold;
(3) bbox correction: we calculate the weighted average between selected samples and the GT box to obtain the corrected GT box. 

\textbf{Quality-aware bag construction.} Our method simultaneously considers the classification probability and the localization score by exploiting the Quality-Aware Learning Strategy (QLS) \cite{zhang2023drone} to form positive proposal bags. The main idea of QLS is to select samples with a higher Quality-Aware Factor (QAF). The QAF of a sample $i$ is defined as:
\begin{equation}
\begin{split}
&\mathrm{QAF}(i)=\max_j(\mathds{1}_S(i)\cdot(\mathrm{HLQ}(i,j)^\alpha\cdot\mathrm{prob}(i,c_j)^{1-\alpha})),\\
&\mathrm{HLQ}(i,j)=\max(\mathrm{SIWD}_a(i,j),\mathrm{SIWD}_p(i,j)),
\end{split}
\end{equation}
where $\mathds{1}_S(i)$ is the spatial prior; $j$ is the index of $j$th GT; $\mathrm{prob}\in[0,1]$ denotes the classification probability; $\alpha\in[0,1]$ controls the weight between classification and localization. HLQ is the Hybrid Location Quality and SIWD is the Scale-Invariant Wasserstein Distance, both of which are proposed in QLS \cite{zhang2023drone} used to better measure the localization quality of a tiny sample. $\mathrm{SIWD}_a(i,j)$ denotes the SIWD value between the anchor of sample $i$ and the GT $j$ while the one with subscript $p$ is used between prediction and GTs. 
In addition, since we are tackling the position shift problem, especially for tiny objects, we adopt QLS \cite{zhang2023drone} for label assignment in the CBC-Head and the final detection head.

\textbf{Batch adaptive threshold and top-1 sample selection.}
In the constructed bag, the proposals have high localization scores with the current GT. However, in our problem setting, the current GT is not accurate, \ie, shifts away from the real object. If still considering the localization score, the selected samples will strongly resemble the current GT and thus lack the correction effect.
Therefore, we consider the classification probability as the score of the sample, which is given by:
\begin{equation}
    \mathrm{score}_i=c_i^*\sigma(p^s_i),
\end{equation}
where $\sigma(\cdot)$ denotes the sigmoid function, $p_i^s$ is the output of the classification branch.

We design a batch adaptive threshold sample selection strategy to prevent low-quality samples from deteriorating the GT box. For example, predictions made under poor illumination conditions are likely to be inaccurate. After bag construction, we calculate a statistic score threshold within a mini-batch:
\begin{equation}
    \mathrm{thr}=\mathrm{Mean}(\{\mathrm{score}_i\})+\mathrm{Std}(\{\mathrm{score}_i\}),
\end{equation}
where $\mathrm{Mean}(\cdot)$ and $\mathrm{Std}(\cdot)$ denote obtaining the mean and standard variation value of a set of numbers; $\{\mathrm{score}_i\}$ is a set, containing the scores of samples in a batch.
For each GT, we select samples with $\mathrm{score}_i$ higher than the batch adaptive threshold. If there is no sample with a higher score, then this GT will not be corrected. 
We further choose the sample with the highest $\mathrm{score}_i$ from the bag after adaptive threshold selection to proceed with bounding box correction.

\textbf{Progressive bounding box center correction.} 
In the context of the prominent position shift problem, we mainly tackle the offset between shifted boxes and real objects. Therefore, we focus on offset correction and leave the scale of the box untouched. We correct the offset by adjusting the center of the box and preserving its height and width. Let $b^{s*}_k=(x_k,y_k,w,h)$ be one of the sensed GT boxes in the $k$th epoch, where $\boldsymbol{p}^*_k=(x_k,y_k)$ is the coordinate of the center and $(w,h)$ denotes the width and height of the box.
The sensed GT boxes are initially set to reference GT boxes, $\mathcal{B}^{s*}_0=\mathcal{B}^{r*}$. The center correction function for $b^{s*}_k$ in $\mathcal{B}^{s*}_{k+1},k\ge0$ is given by: 
\begin{equation}
\boldsymbol{p}^*_{k+1}=\beta \boldsymbol{p} + (1 - \beta)\boldsymbol{p}^*_k,
\end{equation}
where $\boldsymbol{p}$ is the center of the selected sample and $\beta$ is a progressive scalar. We scale down the correction ratio in the early stage of training to prevent the unstable predictions from deteriorating the GT box. $\beta$ is given by:
\begin{equation}
    \beta = \begin{cases}\frac{k}{\left[\gamma\cdot E\right] - 1}, &\text{if }k<\left[\gamma\cdot E\right] \\ 1, &\text{otherwise.}\end{cases}
\end{equation}
where $k\in\{0,1,...,E-1\}$ and $E$ denote the current epoch and the maximum epoch, respectively, symbol $\left[\cdot\right]$ denotes the rounding-to-integer operation and $\gamma\in[0,1]$ is used to adjust the progressive increase range. 

Finally, the total loss in CBC-Head is formulated by:
\begin{equation}
    \mathcal{L}_{cbc}=\frac{1}{2}(\mathcal{L}_{cls}^r+\mathcal{L}_{cls}^s)+\frac{1}{2}(\mathcal{L}_{reg}^r+\mathcal{L}_{reg}^s).
\end{equation}
CBC-Head is adopted only in the training stage and does not influence the inference speed. With the auxiliary supervision of $\mathcal{L}_{cbc}$, the network learns to extract more representative features for both modalities, hence improving the representation of fused RGBT information.

\subsection{Shifted Window-based Cascaded Alignment}
\label{sec:swca}
Another issue caused by the prominent position shift problem is feature map mismatch, which means the identical index of reference and sensed features may respond to discrepant receptive fields. Directly fusing spatially unmatched features element-wisely could lead to confusion. A typical workaround is to adaptively learn sampling positions through a convolution layer~\cite{zhou2020improving,chen2024weakly}. However, the position shift in aerial RGBT image pairs is severe in that the shift is even greater than the object size. The receptive field of a convolution layer can be limited to capture shifted cross-modal semantic relations. More seriously, it usually lacks ground-truth supervision for precise offset prediction. Therefore, RGBT feature alignment under prominent position shift is non-trivial. 

We propose Shifted Window-based Cascaded Alignment (SWCA) for this problem. First, to capture the large shift between modalities, we propose to discover aligned semantic relations by building long-term dependencies between spatially unaligned features. In addition, we utilize the shifted window design in \cite{liu2021swin} to ensure efficiency and achieve cross-modal \& cross-window connection. Furthermore, considering that the objects are so small that their features are fragile against feature deformation, we devise a two-stage cascaded scheme to align the sensed feature delicately.

\begin{figure}[h]
  \centering
  \includegraphics[width=0.49\linewidth]{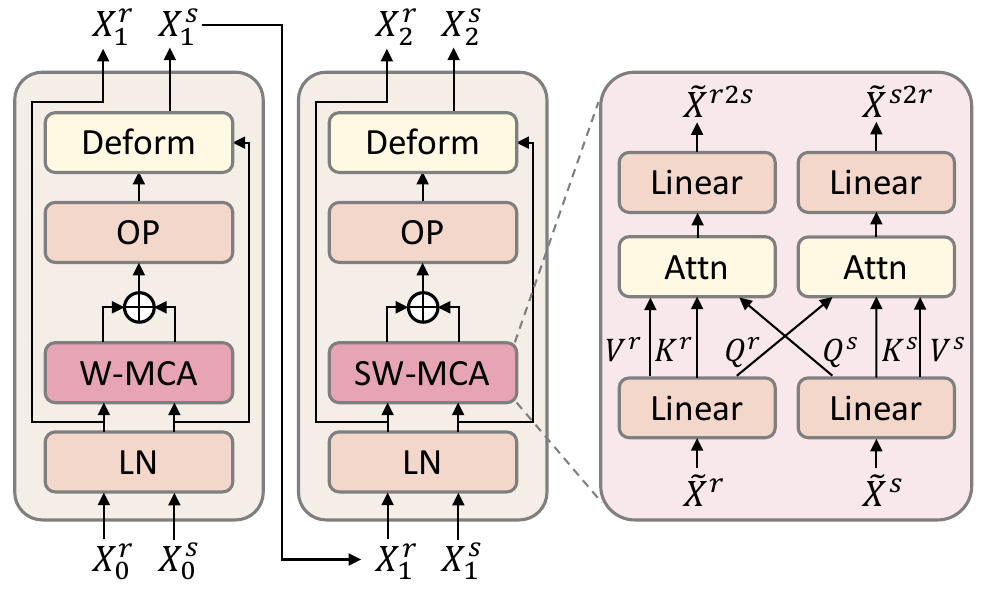}
  \caption{Two successive SWCA Transformer blocks. OP denotes the offset predictor. The cross-attention mechanism in WCA and SWCA is the same except for the window split.
  }
  \label{fig:swca}
\end{figure}

The structure of SWCA is shown in Fig.~\ref{fig:overall}. Given the backbone output feature pair $F^r,F^s\in\mathbb{R}^{B\times C\times H\times W}$, we first flatten these feature maps into embeddings $X^r_0,X^s_0\in\mathbb{R}^{B\times HW\times C}$. $B,C,H,W$ denotes batch size, number of channels, height, and width, respectively. Then, the embedding pair is fed into two successive SWCA blocks, in which the cross-attention based offset prediction and sensed feature deformation are operated. Fig.~\ref{fig:swca} depicts the structure of SWCA blocks and the cross-attention mechanism. The SWCA Transformer consists of two-layer normalization (LN) operations, a cross-attention mechanism, and an offset prediction layer. 

Let $\tilde{X}_r,\tilde{X}_s\in\mathbb{R}^{N_w\times HW\times C}$ denotes embedding $X_r,X_s$ after the LN and window partitioning operation. $N_w$ denotes the number of windows, and $W$ denotes the size of the window. First, $\tilde{X}_r$ is projected by a linear layer to obtain $Q^r,K^r,V^r$. Likewise, we also obtain $Q^s,K^s,V^s$. Then we calculate the cross-attention, which can be formulated as:
\begin{equation}
\begin{aligned}
    A^{r2s}=\mathrm{Softmax}(Q^sK^r/\sqrt{d_k}+P^r),\quad
    A^{s2r}=\mathrm{Softmax}(Q^rK^s/\sqrt{d_k}+P^s),
\end{aligned}
\end{equation}
where $d_k$ is the dimension of $K_r$ and $P^r,P^s$ denotes the relative position bias.
With $A^r,A^s$, we obtain the semantically aligned cross-modality embeddings:
\begin{equation}
    \tilde{X}^{r2s}=A^{r2s}V^r,\quad
    \tilde{X}^{s2r}=A^{s2r}V^s.
\end{equation}
Then we use the cross-modality embedding pair to predict offset $\Delta \boldsymbol{p}_i$. We formulate this procedure as:
\begin{equation}
    \Delta \boldsymbol{p}_i=\mathrm{OP}(\tilde{X}^{r2s}\oplus\tilde{X}^{s2r}),
\end{equation}
where $\mathrm{OP}(\cdot)$ is the offset predictor, a simple linear layer. $\Delta \boldsymbol{p}_i \in\mathbb{R}^{B\times HW\times 2}$ indicates the offset in the $x$ and $y$ direction for every feature point.

Then we restore $\Delta\boldsymbol{p}_i$ to the shape $(B\times H\times W\times 2)$, $X^r_i$ and $X^s_i$ to $F^r_i\in\mathbb R^{B\times H\times W\times C}$ and $F^s_i\in\mathbb R^{B\times H\times W\times C}$ after window reverse. We generate a sampling grid $\mathcal{T}_i(G)$ by applying $\Delta\boldsymbol{p}_i$ to the regular grid $G$. Finally, we sample the sensed feature $F^s_i$ by grid $\mathcal{T}_i(G)$, obtaining deformed sensed feature $\hat{F}^s_i$, which is aligned with the reference feature.

\section{Experiment}
\label{sec:exp}

\subsection{Datasets}
\textbf{RGBTDronePerson. }Proposed by Zhang \etal ~\cite{zhang2023drone}, it is a drone-based RGBT tiny person detection dataset, containing 6,125 pairs of RGBT images. The dataset focuses on person-centric targets, namely, \textit{person}, \textit{rider}, and \textit{crowd}. The average size of the objects is only 11.7 pixels. The prominent position shift problem is first put forward by RGBTDronePerson, yet there is a lack of study and efficient solutions to this issue. This dataset is only annotated on the thermal modality. In this work, to quantitatively evaluate the bbox correction, we annotate the first 1,200 visible images in the training set.

\textbf{DroneVehicle. }It is a drone-based RGBT vehicle detection dataset by Sun \etal~\cite{sun2022drone}, containing 5 categories of vehicles, namely, \textit{car}, \textit{truck}, \textit{bus}, \textit{van}, \textit{freight car}. Different from RGBTDronePerson, it is an oriented object detection dataset. Moreover, this dataset comes with both visible annotations and thermal annotations, helping us study the position shift problem more thoroughly. To further validate the advantage of our method against large position shifts, we select a shift subset from DroneVehicle.

\subsection{Evaluation Metrics}

\textbf{mAP.} It is used to evaluate the detection performance. Following \cite{zhang2023drone} and \cite{sun2022drone}, we take the mAP at the IoU threshold of 0.5 as the main metric on RGBTDronePerson and DroneVehicle datasets.

\textbf{aSim.} In this work, we introduce a new metric to measure the similarity between two pairs of box annotations. Given a pair of visible and thermal images, we assume that both are annotated with bounding boxes. We conduct one-to-one matching between the two sets of boxes by the Hungarian algorithm and compute the box similarity between matched boxes. Then we average the similarity scores of all boxes and all image pairs in the training set, obtaining aSim. 
This process can be formulated as:
\begin{equation}
    \mathrm{aSim}=\sum_i^N\sum_j^M \mathrm{Sim}(b_{ij}^r,\tilde{b}_{ij}^s),
\end{equation}
where $N$ and $M$ denote the number of image pairs and boxes in one pair, respectively; $b_{ij}^r$ denotes the $j$th reference box in the $i$th image; $\tilde{b}_{ij}^s$ denotes the matched sensed box; $\mathrm{Sim}$ denotes box similarity (IoU, GIoU~\cite{rezatofighi2019generalized}, SIWD~\cite{zhang2023drone}, etc.). For RGBTDronePerson, we use SIWD as the box similarity metric. For DroneVehicle, we use IoU since SIWD is designed for horizontal boxes. In the subsequent experiments, we utilize aSim to quantitatively evaluate the object position shift between modalities and to evaluate our box correction results.

\subsection{Implementation Details}
We implement our method under MMDetection V2~\cite{mmdetection}, MMRotate~\cite{zhou2022mmrotate}, and Pytorch~\cite{pytorch} framework. Experiments are conducted on an Nvidia GTX 4090 GPU. We build our detector based on the ATSS detector~\cite{zhang2020atss} and use Swin-S~\cite{liu2021swin} as the backbone. Our baseline is ATSS with QLS~\cite{zhang2023drone}. Data augmentation only involves horizontal random flipping. In oriented object detection, we simply use ATSS (OBB) with QLS as the base detector. We substitute the SIWD in QLS with IoU since SIWD is not implemented for oriented object detection. Data augmentation includes horizontal, vertical, and diagonal random flipping. Detectors are trained for 24 epochs and optimized by AdamW with an initial learning rate of $5\times 10^{-5}$, (0.9, 0.999) betas, and 0.05 weight decay. We pre-train a baseline model on the reference modality and take the weight of its detection head for initializing the CBC-Head ensemble. The smooth coefficient parameter of the EMA is 0.9997. For RGBTDronePerson and DroneVehicle shift sub-set, we only utilize the top-1 sample selection, and the parameter for bounding box correction $\gamma$ is 0.5. For the DroneVehicle full-set, we apply batch adaptive threshold and top-1 sample selection together and $\gamma$ is set to 1.0.

\subsection{Overall Performance on RGBTDronePerson}
We compare our method with state-of-the-art RGBT detection methods, including HRFuser~\cite{broedermann2023hrfuser}, TINet~\cite{zhang2023tinet}, ICAFusion~\cite{shen2024icafusion}, C$^2$Former~\cite{yuan2024c}, QFDet~\cite{zhang2023drone} on RGBTDronePerson. Our method achieves an $\mathrm{mAP}_{50}$ of 43.55. In particular, our method surpasses the second-best method by 3.70 for the majority category \textit{person}. For category \textit{rider}, which is with large position shifts, our method achieves the highest $\mathrm{mAP}_{50}$ of 53.62. We also compare with C$^2$Former on RGBTDronePerson, which also aims at miscalibration between RGB and the thermal modality. As it is based on an oriented detector, we take the C$^2$Former backbone and put it into an ATSS detector. C$^2$Former performs well on the scarce category \textit{crowd} but lags on \textit{person} and \textit{rider}. The category \textit{ crowd} accounts for only 10\% of the total instances and presents large intraclass variations since a number of persons can form a crowd in various ways. Therefore, the accuracy on \textit{crowd} is relatively low and fluctuates. However, our method does effectively improve the performance on tiny objects (\textit{person}) and high-speed objects (\textit{rider}), which proves again that the prominent position shift makes a serious impact on these objects.

\begin{table}[htbp]
  \caption{Performance comparison between state-of-the-art RGBT detection methods on RGBTDronePerson. C$^2$Former$^*$ means we implement C$^2$Former into horizontal object detection based on the ATSS detector.
  }
  \label{tab:dp}
  \centering
  \begin{tabular}{ccccc}
    \toprule
    Methods & $\mathrm{mAP}_{50}$ & person & rider & crowd \\ 
    \midrule
    HRFuser~\cite{broedermann2023hrfuser} & 22.23 & 16.25 & 24.82 & 25.64  \\
    TINet~\cite{zhang2023tinet} & 28.30 & 15.21 & 43.38 & 26.30 \\
    ICAFusion~\cite{shen2024icafusion} & 28.56 & 38.40 & 19.30 & 28.00  \\
    C$^2$Former$^*$~\cite{yuan2024c} & 37.71 & 37.35 & 45.63 & \textbf{30.16} \\
    QFDet~\cite{zhang2023drone} & 42.08 & 46.06 & 50.26 & 29.91\\
    Ours & \textbf{43.55} & \textbf{49.76} & \textbf{52.62} & 28.27 \\
  \bottomrule
  \end{tabular}
\end{table}

\begin{figure*}[htbp]
  \centering
  \includegraphics[width=0.99\linewidth]{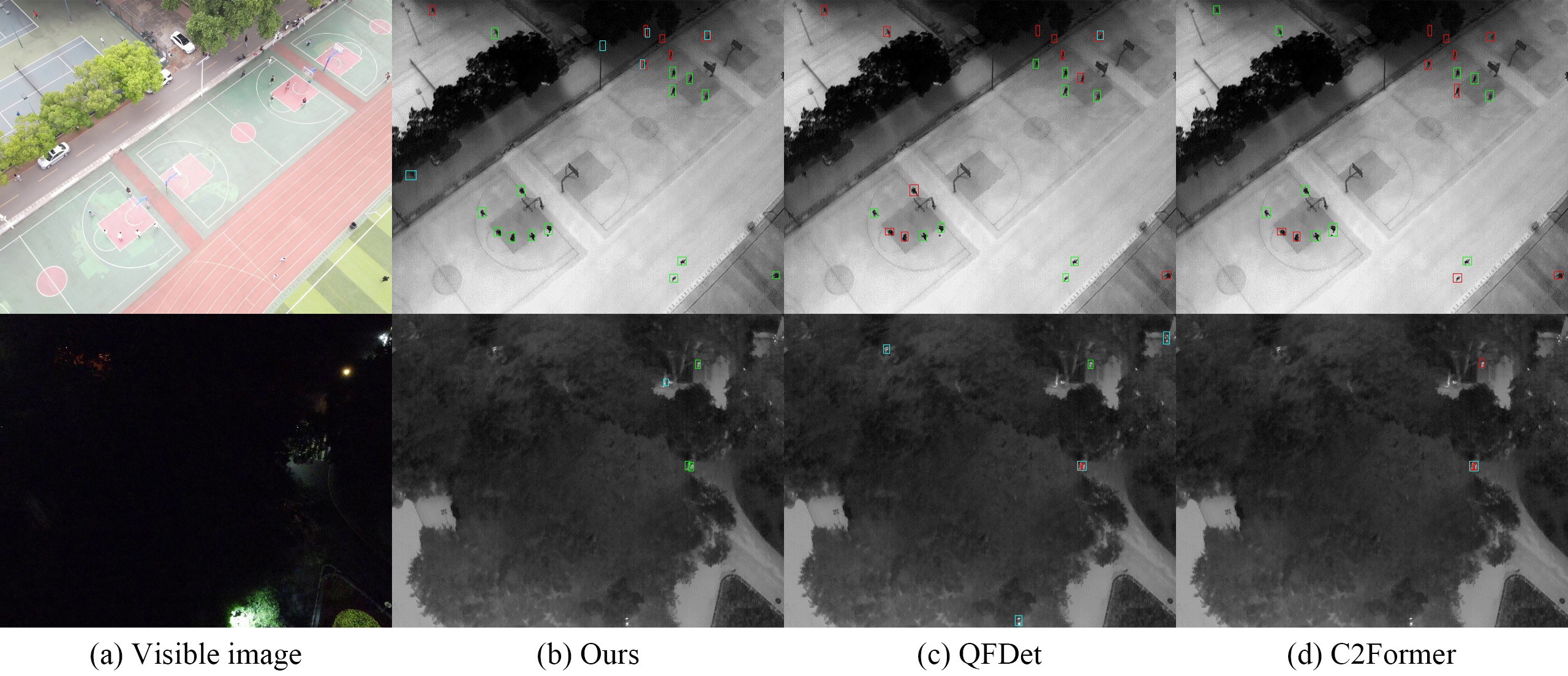}
  \caption{Detection results on RGBTDronePerson. Green boxes denote true positives; red boxes denote false negatives; blue boxes denote false positives. (a) Visible images. (b) Thermal images and detection results of our method. (c) Detection results of the QFDet. (d) Detection results of C$^2$Former.
  }
  \label{fig:dpd}
\end{figure*}

We compare the visualized detection results among our method, QFDet, and C$^2$Former in Fig.~\ref{fig:dpd}. The first row shows daytime scenes and the last row shows nighttime scenes. From the daytime scene, we can see that the ground obtains a higher temperature than human bodies, and thus, the persons present black in the thermal modality, unlike normal scenes. QFDet and C$^2$Former fail to detect multiple objects in this scene, while our detector is more robust. This is because the other two methods tend to overfit in the normal scenarios of the thermal (reference) modality and overlook the visible (sensed) modality. At nighttime, our detector produces fewer false alarms than the other detectors.

\begin{figure*}[htbp]
  \centering
  \includegraphics[width=0.89\linewidth]{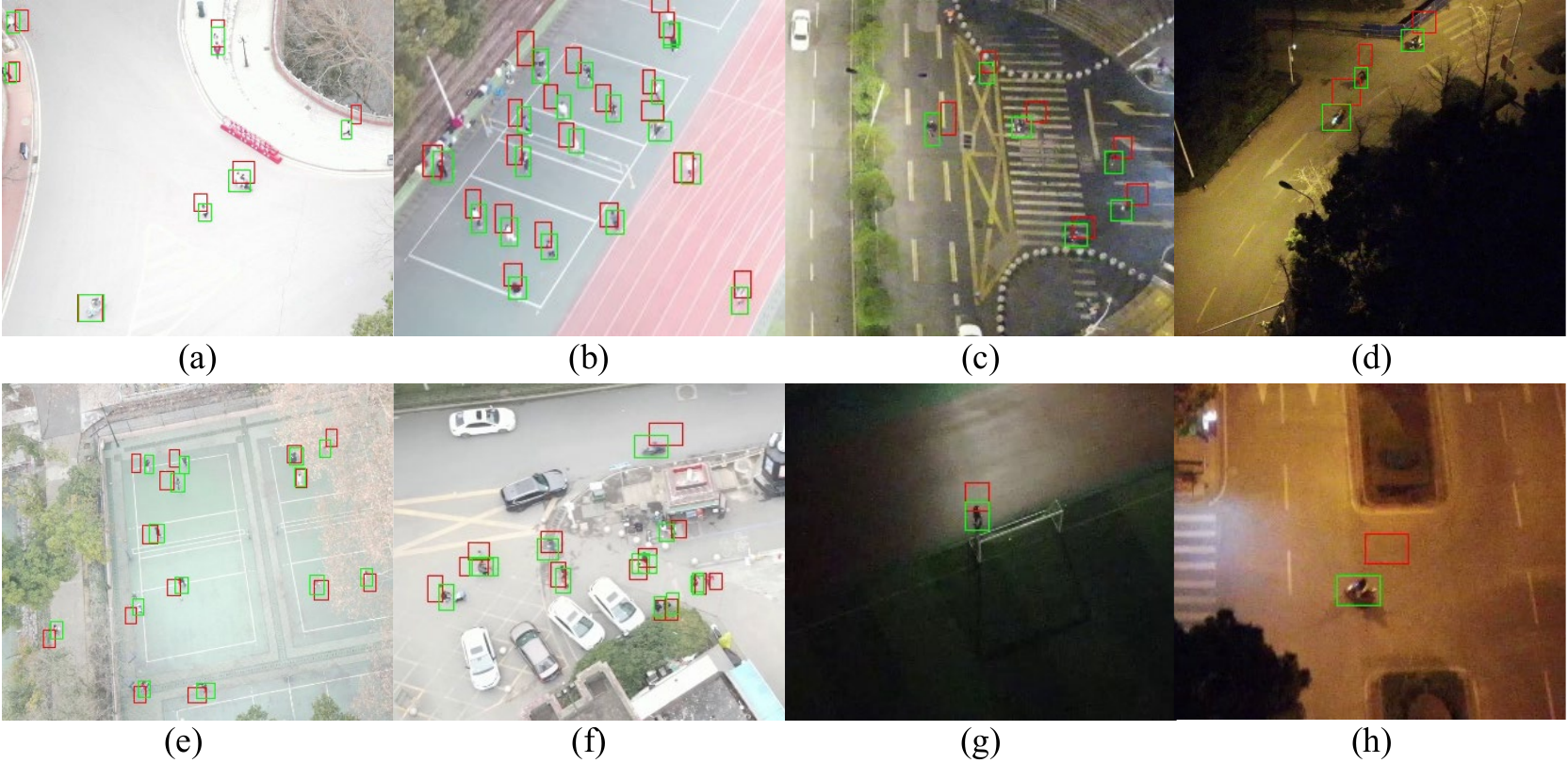}
  \caption{Correction results on RGBTDronePerson. Every sub-figure is a zoomed-in area of one image to show the corrected tiny objects clearly. Red boxes denote shifted GTs and green boxes denote corrected GTs. (a)-(h) are eight examples from different images.
  }
  \label{fig:dpc}
\end{figure*}

\textbf{Correction performance.} We visualize the sensed GT correction results in Fig.~\ref{fig:dpc} to showcase the correction performance of the CBC module. CBC can adaptively correct the bboxes. Especially from Fig.~\ref{fig:dpc} (a) and (f) we can see that for some objects their original bboxes are accurate so CBC preserves their bboxes; for some objects, the original bboxes shift far away from them, CBC can still correct the boxes to their accurate positions. Fig.~\ref{fig:dpc} (c),(d), (g), and (h) demonstrate the performance in nighttime scenarios. In the presence of the domain gap between daytime and nighttime, CBC manages to correct shifted boxes day and night. We attribute it to the cross-modality Mean Teacher framework in CBC, which effectively leverages the thermal information. However, we must admit that without any artificial light, CBC may not be able to sense the real positions of objects, and the same goes for human eyes. We also evaluate the improvement on aSim using daytime testing images with annotations. The aSim between the original thermal boxes and visible boxes is only 22.57. With our cross-modality bbox correction, the aSim between corrected visible boxes and real visible boxes is improved to 48.09 (25.52$\uparrow$).

\subsection{Overall Performance on DroneVehicle}
DroneVehicle comes with annotations for the visible and the thermal modality. Therefore, we investigate the ability to correct cross-modality bbox from two settings: (1) thermal reference (TR) and (2) visible reference (VR). TR means we take the annotations for thermal modality as the reference GT and use the annotations for visible modality only for corrected box evaluation. For VR, it is vice versa. Unless stated otherwise, the results are given under the TR setting in this subsection. We will discuss the performance under the VR setting in the Discussion subsection.

We compare with state-of-the-art RGBT detectors on the DroneVehicle val set and test set. Results are shown in Tab.~\ref{tab:dv}. Our method achieves the highest mAPs both on the val set and test set, which are 77.6 and 76.5 points. With sufficient cross-modality complementary information exploitation, our method succeeds in detecting difficult categories like \textit{truck}, \textit{van}, and \textit{freight car}. AR-CNN~\cite{zhang2019weakly}, TSFADet~\cite{yuan2022translation}, CAGTDet~\cite{yuan2024improving}, and C$^2$Former~\cite{yuan2024c} are designed for weak misalignment, in which AR-CNN, TSFADet, and CAGTDet require annotations for both modalities to achieve region alignment. However, our method utilizes only reference annotations and explicitly produces sensed annotations to improve multi-modal representation learning and benefit adaptive feature alignment. The category \textit{car} is the dominant category in DroneVehicle, accounting for 85.5\% of the total instances in the training set. Detectors intend to bias towards \textit{car} and obtain better accuracy on \textit{car}. In our method, the CBC-Head provides auxiliary supervision on classification and thus alleviates the bias towards \textit{car}, effectively elevating the $\mathrm{mAP}$ on the minor sacrifice of \textit{car}.
\begin{table}[htbp]
  \caption{Performance comparison between state-of-the-art RGBT detection methods on DroneVehicle.
  }
  \label{tab:dv}
  \centering
  \begin{tabular}{ccccccc}
    \toprule
    \multicolumn{5}{l}{DroneVehicle val set.} \\ \hline
    Methods & $\mathrm{mAP}$ & car & truck & bus & van & freight car \\
    \hline
    Halfway Fusion (OBB)~\cite{liu2016multispectral} & 68.2 & 89.9 & 60.3 & 89.0 & 46.3 & 55.5 \\
    CIAN (OBB)~\cite{zhang2019cross} & 70.2 & 90.0 & 62.5 & 88.9 & 49.6 & 60.2 \\
    AR-CNN (OBB)~\cite{zhang2019weakly} & 71.6 & 90.1 & 64.8 & 89.4 & 51.5 & 62.1 \\
    MBNet (OBB)~\cite{zhou2020improving} & 71.9 & 90.1 & 64.4 & 88.8 & 53.6 & 62.4 \\
    TSFADet~\cite{yuan2022translation} & 73.1 & 90.0 & 67.9 & 89.8 & 54.0 & 63.7 \\
    C$^2$Former~\cite{yuan2024c} & 74.2 & 90.2 & 68.3 & 89.8 & 58.5 & 64.4 \\
    CAGTDet~\cite{yuan2024improving} & 74.6 & \textbf{90.8} & 69.7 & \textbf{90.5} & 55.6 & 66.3 \\
    Ours & \textbf{77.6} & 89.8 & \textbf{75.5} & 89.2 & \textbf{63.0} & \textbf{70.6} \\
    \hline
    \multicolumn{5}{l}{DroneVehicle test set.} \\ \hline
    UA-CMDet~\cite{sun2022drone} & 64.0 & 87.5 & 60.7 & 87.1 & 38.0 & 46.8 \\ 
    TSFADet~\cite{yuan2022translation} & 70.4 & 89.2 & 72.0 & 88.1 & 48.8 & 54.2 \\
    C$^2$Former~\cite{yuan2024c} & 73.3 & 89.1 & 66.5 & 88.1 & 57.5 & \textbf{65.5} \\
    CALNet~\cite{he2023multispectral} & 73.1 & \textbf{90.2} & 73.8 & 88.7 & 58.5 & 60.9 \\
    Ours & \textbf{76.5} & 87.8 & \textbf{76.0} & \textbf{89.2} & \textbf{65.0} & 64.2 \\
  \bottomrule
  \end{tabular}
\end{table}

\begin{table}[htbp]
  \caption{Effect validation on DroneVehicle shift subset.
  }
  \label{tab:dvs}
  \centering
  \begin{tabular}{ccccccc}
    \toprule
    Methods & $\mathrm{mAP}$ & car & truck & bus & van & freight car \\
    \midrule
    C$^2$Former~\cite{yuan2024c} & 62.0 & 88.1 & 61.0 & 85.4 & 43.9 & 31.4 \\
    Baseline & 66.6 & 88.0 & 62.4 & 86.6 & 50.4 & 45.5 \\
    Ours & \textbf{70.6} & \textbf{88.7} & \textbf{67.1} & \textbf{88.0} & \textbf{60.9} & \textbf{48.4} \\
   \bottomrule
  \end{tabular}
\end{table}

\begin{figure*}[htbp]
  \centering
  \includegraphics[width=0.89\linewidth]{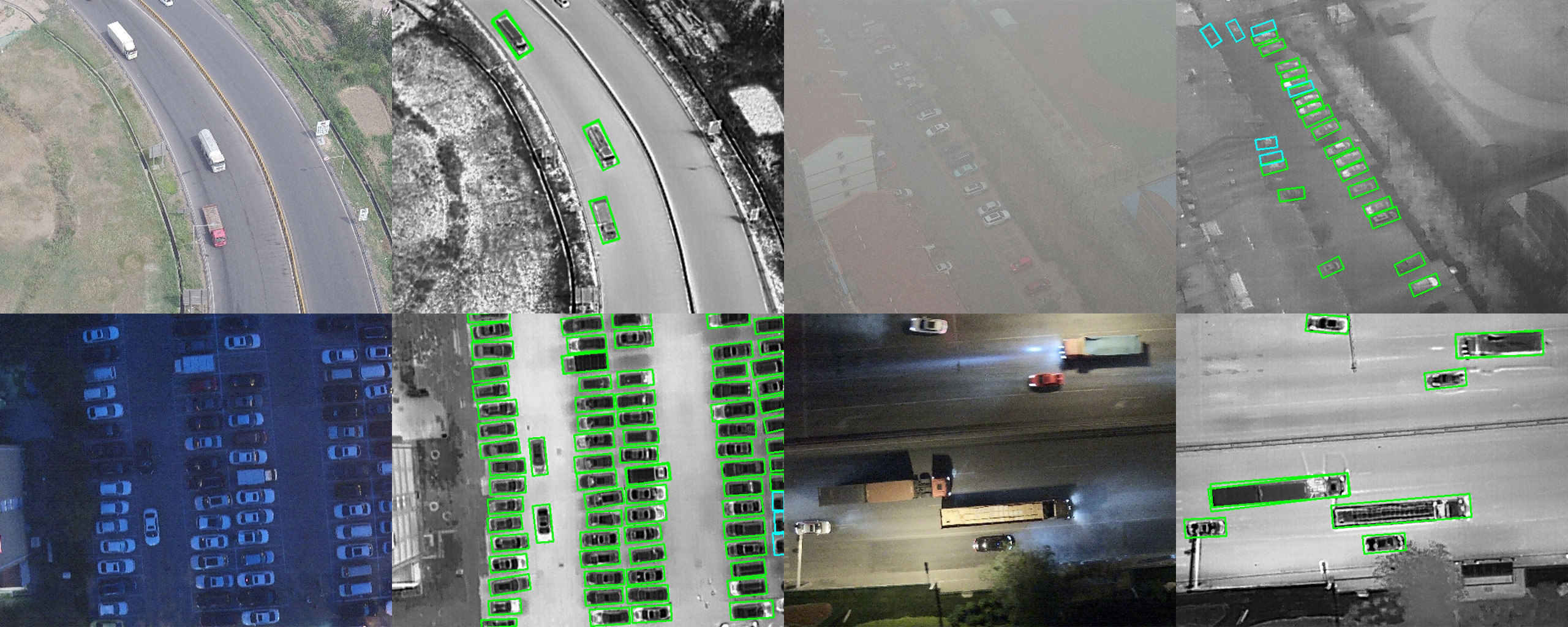}
  \caption{Detection results of our method on DroneVehicle. Green boxes denote true positives; blue boxes denote false positives. 
  }
  \label{fig:dvd}
\end{figure*}

\textbf{Shift subset.} We evaluate the aSim between visible GT boxes and thermal GT boxes. We calculate the aSim of every image pair, set the mean value of aSim subtract its standard variation value as a threshold, and select those image pairs with aSim lower than the threshold into the shift subset. 
The aSim of the shift training sub-set is only 66.38\% while the aSim of the full set is 83.47\%. The shift subset contains 6,185 pairs of images for training, accounting for 34.4\% of the full train set. The shift test sub-set contains 3,157 pairs and the aSim is 66.62\%. To validate the effectiveness of our method under the prominent position shift scenario, we compare C$^2$Former, baseline, and our method on the shift subset. The results are shown in Tab.~\ref{tab:dvs}. C$^2$Former also faces the misalignment problem with GTs of one modality. Our method surpasses C$^2$Former by 8.6 points and improves the baseline by 4.0 points on mAP. C$^2$Former degrades 11.1 points from DroneVehicle full-set to shift sub-set (73.1 to 62.0), while our method only declines 5.9 points (76.5 to 70.6). It well demonstrates the ability of our method against the prominent position shift problem. Fig.~\ref{fig:dvd} shows some detection results on DroneVehicle. For densely arranged and fast-moving scenarios, our method achieves robust detection performance.


\begin{figure*}[htbp]
  \centering
  \includegraphics[width=0.89\linewidth]{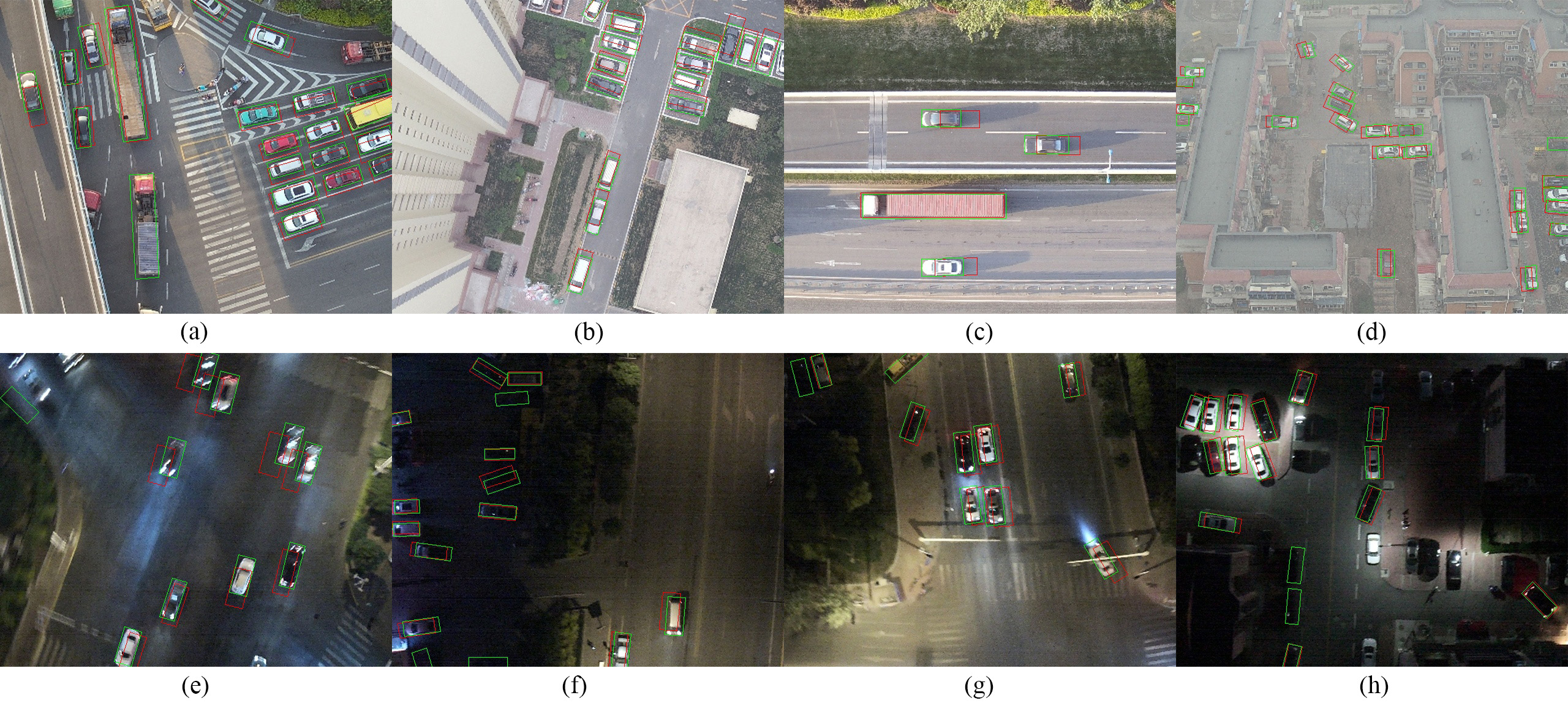}
  \caption{Correction results on DroneVehicle. Red boxes denote shifted GTs and green boxes denote corrected GTs. (a)-(h) are eight examples from different images.
  }
  \label{fig:dvc}
\end{figure*}

\textbf{Correction performance. }We visualize the sensed GT correction results in Fig.~\ref{fig:dvc}. The first row gives the results in the daytime and the second row shows the nighttime. Fig.~\ref{fig:dvc} (a) and (c) show that when objects are moving at a higher speed, the position shifts are more prominent. Fig.~\ref{fig:dvc} (b) and (d) demonstrate that the position shifts will also present due to FoV discrepancy. Fig.~\ref{fig:dvc} (e) shows the sudden rotation of the drone will also cause severe position shifts. Our cross-modality bbox correction manages to adaptively produce finer annotations under various occasions, including unsatisfying illumination conditions as shown in Fig.~\ref{fig:dvc} (f), (g), and (h). We also quantitatively evaluate the correction performance. The aSim between corrected sensed GT and real sensed GT is 76.42, which is 10.04 points higher than the original (66.38).

\subsection{Ablation Studies}
To showcase the design of CBC and SWCA, we perform detailed ablation studies on the design choices and performance comparison. Tab.~\ref{tab:mtbcswa} gives the ablation studies for CBC and SWCA on RGBTDronePerson. We improve the baseline by 2.16 points on $\rm{mAP}_{50}$ with CBC alone. The majority class \textit{person} improves by 3.53 points. Class \textit{rider}, which presents a bigger position shift, improves by 2.47 points. The SWCA module improves the baseline by 1.42 points on $\rm{mAP}_{50}$ and enhances the tiny \textit{person} by 3.27 points. SWCA alone does not improve \textit{rider}, which could be that the large position shifts between modalities hinder the RGBT representation learning and further invalidate feature alignment. CBC and SWCA share a mutual benefit. When applied together, our detector achieves 43.55 points on $\mathrm{mAP}_{50}$, improves the baseline on \textit{person} and \textit{rider} by 4.66 points and 3.85 points respectively. Category \textit{crowd} accounts for only 10\% of the total instances and it presents large intra-class variations. Therefore, the accuracy on \textit{crowd} is relatively low and fluctuates.

\begin{table}[htbp]
  \caption{Ablation studies for two main designs on RGBTDronePerson.
  }
  \label{tab:mtbcswa}
  \centering
  \begin{tabular}{cccccc}
    \toprule
    CBC & SWCA & $\mathrm{mAP}_{50}$ & person & rider & crowd \\
    \midrule
     & & 40.90 & 45.30 & 48.77 & 28.65 \\
    \ding{51} &  & 43.06 & 48.83 & 51.24 & 29.11 \\
     & \ding{51} & 42.52 & 48.57 & 48.75 & \textbf{30.25} \\
    \ding{51} & \ding{51} & \textbf{43.55} & \textbf{49.76} & \textbf{52.62} & 28.27 \\
  \bottomrule
  \end{tabular}
\end{table}

\textbf{Mean Teacher framework in CBC.} We incorporate the Mean Teacher (MT) framework into our RGB-T sensed-reference problem setting to alleviate the modality discrepancy and improve the sensed GT correction. We validate the effectiveness of MT on RGBTDronePerson, as shown in Tab.~\ref{tab:cbc_mt}. We can see that the correction and detection accuracy both benefit from MT.

\begin{table}[htbp]
  \caption{Experimental results validating Mean Teacher framework in CBC on RGBTDronePerson. 
  }
  \label{tab:cbc_mt}
  \centering
  \begin{tabular}{c|c|cccc}
    \toprule
    Method & aSim & $\mathrm{mAP}_{50}$ & person & rider & crowd \\
    \midrule
    w/ MT & \textbf{48.09} & \textbf{43.06} & 48.83 & \textbf{51.24} & \textbf{29.11} \\
    w/o MT & 46.67 & 42.33 & \textbf{49.07} & 49.99 & 27.94 \\
  \bottomrule
  \end{tabular}
\end{table}

\textbf{Choice of $\gamma$ in CBC.} We use a progressive scalar $\beta$ to control the correction amplitude and further define the curve of $\beta$ by $\gamma$. $\gamma$ controls slop when $\beta$ linearly grows to its max value. In Tab.~\ref{tab:bg} we conduct detailed studies on the value of $\gamma$. When $\gamma=0.0$, the network takes the predicted boxes as corrected GTs from the beginning, which inevitably will introduce unstable predictions. The drop of aSim is also observed in the last few epochs. When $\gamma=0.5$, the network takes the average of the predicted box and previous GT box as the corrected box in the first half of training epochs, yielding the best aSim and $\rm{mAP}_{50}$ values. When $\gamma=1.0$, the network relies much on previous inaccurate GT boxes during the training stage and thus does not produce better GT boxes for the sensed modality. Therefore, we set $\gamma$ to 0.5 for RGBTDronePerson.

\begin{table}[htbp]
  \caption{Ablation studies on $\gamma$ in CBC.
  }
  \label{tab:bg}
  \centering
  \begin{tabular}{c|c|cccc}
    \toprule
    Method & aSim & $\mathrm{mAP}_{50}$ & person & rider & crowd \\
    \midrule
    0.0 & 40.17 & 42.42 & 49.19 & 49.69 & 28.39 \\
    0.5 & \textbf{48.09} & \textbf{43.06} & \textbf{49.75} & 50.74 & \textbf{28.68} \\
    1.0 & 28.50 & 42.51 & 49.25 & \textbf{51.00} & 27.07 \\
    \bottomrule
  \end{tabular}
\end{table}

\textbf{Effectiveness of box correction. }We ablate the box correction procedure in the CBC head ensemble. Specifically, we do not correct the sensed GT boxes. Instead, we simply use the reference GT boxes for the senesd modality. The experimental results are shown in Tab.~\ref{tab:cbc_bc}. The first row gives the results of the CBC head ensemble with box correction and the second row gives the results of the CBC head ensemble not correcting the sensed GT box. The aSim between reference and sensed GTs is only 22.57. With box correction, we elevate it by 25.52 to 48.09. As for $\rm{mAP}$, although ``w/o BC'' simply uses reference GTs for the sensed modality, the additional supervision provided by CBC Head ensemble also helps the detector, improving the baseline by points on $\rm{mAP}_{50}$. With box correction, the $\rm{mAP}_{50}$ further increases to 43.06 points. Notably, class \textit{Rider} reaches 51.24 points on $\rm{AP}_{50}$.

\begin{table}[htbp]
  \caption{Experimental results validating sensed GT box correction in CBC on RGBTDronePerson. 
  }
  \label{tab:cbc_bc}
  \centering
  \begin{tabular}{c|c|cccc}
    \toprule
    Method & aSim & $\mathrm{mAP}_{50}$ & person & rider & crowd \\
    \midrule
    w/ BC & \textbf{48.09} & \textbf{43.06} & \textbf{48.83} & \textbf{51.24} & \textbf{29.11} \\
    w/o BC & 22.57  & 42.06 & 47.83 & 49.53 & 28.82 \\
  \bottomrule
  \end{tabular}
\end{table}

\begin{figure*}[htbp]
  \centering
  \includegraphics[width=0.99\linewidth]{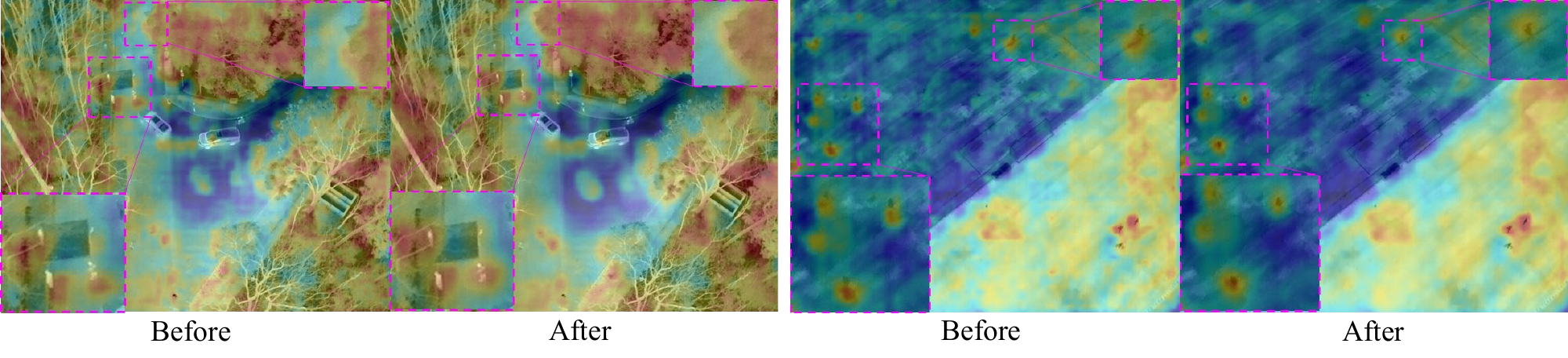}
  \caption{Sensed feature maps before and after SWCA. The base image is the reference image. The dotted circles show the sensed feature map is aligned to the reference image.
  }
  \label{fig:swa}
\end{figure*}

\textbf{Effectiveness of SWCA.} We visualize the sensed feature maps before and after SWCA in Fig.~\ref{fig:swa}. In SWCA, sensed features are deformed to align with the reference features. Therefore, we visualize the sensed feature maps on reference images to show that SWCA effectively aligns the sensed feature maps spatially to the reference modality. The left of Fig.~\ref{fig:swa} depicts one scene, where we can see that the responses for objects are aligned to the real positions in the reference modality, and the silhouette of background trees is aligned with the feature map response. The right shows another scene where the responses for objects are apparently aligned to the real positions in the reference modality.

\textbf{Validation of cross-attention based offset prediction. }One of the common practices to deform the feature map is to predict the feature point offset by one or two convolution layers and deform the map with grid sample. We suppose this common practice is not enough for prominent position shift circumstances and propose to predict offsets based on window cross-attention. We validate this proposal by experiments, as shown in Tab.~\ref{tab:swa_comp}. Convolution-based offset deformation spoils the sensed features and degrades the detection performance while our method effectively aligns the sensed and reference features and improves detection accuracy.

\begin{table}[htbp]
  \caption{Experimental results comparing deformation methods on RGBTDronePerson. ``None'' denotes the baseline model without feature deformation. ``Conv'' denotes convolution-based offset prediction.
  }
  \label{tab:swa_comp}
  \centering
  \begin{tabular}{ccccc}
    \toprule
    Methods & $\mathrm{mAP}_{50}$ & person & rider & crowd \\ 
    \midrule
    None & 40.90 & 45.30 & 48.77 & 28.65 \\
    Conv & 39.69 & 46.44 & 44.88 & 27.76 \\
    Ours & \textbf{42.52} & \textbf{48.57} & \textbf{48.75} & \textbf{30.25} \\
  \bottomrule
  \end{tabular}
\end{table}

\textbf{Number of blocks and shifted windows in SWCA.} Tab.~\ref{tab:swa_block} gives the experimental results regarding the number of blocks in SWCA. One SWCA block is applied with the normal window partition. Two SWCA blocks involve one normal window partition and one shifted window partition, as introduced in Sec.~\ref{sec:meth}. Here the 4-block setting is equal to cascading two 2-block settings. Compared to 2-block, the ability of 1-block SWCA to predict large offsets is weaker without cross-attention in shifted windows, thereby obtaining lower accuracy. Compared to 2-block, not only does 4-block consume more memory, but also degrades the accuracy.
\begin{table}[htbp]
  \caption{Ablation studies on the number of blocks in SWCA on RGBTDronePerson dataset.
  }
  \label{tab:swa_block}
  \centering
  \begin{tabular}{ccccc}
    \toprule
    Number & $\mathrm{mAP}_{50}$ & person & rider & crowd \\
    \midrule
    1 & 42.05 & 48.24 & 47.93 & 29.98 \\
    2 & \textbf{42.52} & \textbf{48.57} & \textbf{48.75} & \textbf{30.25} \\
    4 & 41.72 & 47.64 & 48.18 & 29.34 \\
  \bottomrule
  \end{tabular}
\end{table}


\textbf{Different reference modalities.} In the previous experiments, we take thermal modality as the reference modality. Thermal modality is robust to illumination variation while visible modality degrades at nighttime. What if we take visible modality as the reference modality? How will our cross-modality bbox correction be affected? We discuss these questions by switching the reference modality to visible on the DroneVehicle shift sub-set. The results shown in Tab.~\ref{tab:dvs_vref} indicate that the different reference modality does not impact the detection accuracy. The baseline model supervised by thermal annotations obtains an $\mathrm{mAP}_{50}$ of 66.6 and the one supervised by visible annotations obtains 67.0. Our method improves the baseline by 4 points with thermal annotations and 3.4 points with visible annotations. ``aSim'' shows the cross-modality bbox correction ability of our method. When the reference modality is thermal, we calculate the aSim between corrected GTs and the visible annotated GTs; when the reference modality is visible, vice versa. We can see that the correction ability is robust under different reference modalities. The aSim under VR is a bit higher than that under TR, which could be that the CBC-Head corrects better in darkness with thermal modality than with visible modality.

\begin{table}[htbp]
  \caption{Experimental results on DroneVehicle shift subset with different reference modalities.
  }
  \label{tab:dvs_vref}
  \centering
  \begin{tabular}{c|c|c|cccccc}
    \toprule
    Methods & Reference & aSim & $\mathrm{mAP}$ & car & truck & bus & van & freight car \\
    \midrule
    C$^2$Former & \multirow{3}{*}{Thermal} & - & 62.0 & 88.1 & 61.0 & 85.4 & 43.9 & 31.4 \\
    Baseline & & - & 66.6 & 88.0 & 62.4 & 86.6 & 50.4 & 45.5 \\
    Ours & & 76.4 & \textbf{70.6} & \textbf{88.7} & \textbf{67.1} & \textbf{88.0} & \textbf{60.9} & \textbf{48.4} \\
    \hline
    C$^2$Former & \multirow{3}{*}{Visible} & - & 62.9 & 88.2 & 61.6 & 85.2 & 44.9 & 34.8 \\
    Baseline & & - & 67.0 & 87.2 & 63.6 & 86.8 & 53.5 & 43.8 \\
    Ours & & 77.7 & \textbf{70.4} & \textbf{88.7} & \textbf{67.6} & \textbf{87.9} & \textbf{61.2} & \textbf{47.3} \\
   \bottomrule
  \end{tabular}
\end{table}

\section{Conclusion}
In this paper, we address the unique prominent position shift problem in the context of drone-viewed RGBT object detection, which will cause bounding box confusion and feature map mismatch for the detector. We consider the shifted GT boxes of sensed modality as inaccurate GT boxes and manage to correct those boxes on the fly. We achieve this not only by introducing the MIL scheme but also by taking advantage of modal-shared information with the Mean Teacher framework. Further, we build a Cross-modality Bbox Correction module, which promotes the representation learning of both modalities and further improves the RGBT detection accuracy. As for the feature map mismatch problem, we devise a Shifted Window-based Cascaded Alignment module to achieve delicate feature deformation and alignment between the reference and sensed modalities. Experiments on the RGBTDronePerson dataset and DroneVehicle shift sub-set strongly prove the ability of our method against prominent position shifts and show our method consistently improves the detection accuracy. Our cross-modality bbox correction module automatically provides more accurate annotations for the sensed modality along with the training stage. Apart from the detection task, this technique could help alleviate the cost of multi-modal labeling. In future work, we will study the imbalanced multimodal learning problem, which is coupled with the prominent position shift problem in drone-based RGBT object detection.

\bibliographystyle{plain}  
\bibliography{references}  

\begin{thebibliography}{10}

\bibitem{zhang2019weakly}
Lu~Zhang, Xiangyu Zhu, Xiangyu Chen, Xu~Yang, Zhen Lei, and Zhiyong Liu.
\newblock Weakly aligned cross-modal learning for multispectral pedestrian
  detection.
\newblock In {\em Proceedings of the IEEE/CVF International Conference on
  Computer Vision}, pages 5127--5137, 2019.

\bibitem{kaist}
Soonmin Hwang, Jaesik Park, Namil Kim, Yukyung Choi, and In~So~Kweon.
\newblock Multispectral pedestrian detection: Benchmark dataset and baseline.
\newblock In {\em Proceedings of the IEEE Conference on Computer Vision and
  Pattern Recognition}, pages 1037--1045, 2015.

\bibitem{jia2021llvip}
Xinyu Jia, Chuang Zhu, Minzhen Li, Wenqi Tang, and Wenli Zhou.
\newblock Llvip: A visible-infrared paired dataset for low-light vision.
\newblock In {\em Proceedings of the IEEE/CVF International Conference on
  Computer Vision}, pages 3496--3504, 2021.

\bibitem{zhang2023drone}
Yan Zhang, Chang Xu, Wen Yang, Guangjun He, Huai Yu, Lei Yu, and Gui-Song Xia.
\newblock Drone-based {RGBT} tiny person detection.
\newblock {\em ISPRS Journal of Photogrammetry and Remote Sensing}, 204:61--76,
  2023.

\bibitem{yuan2022translation}
Maoxun Yuan, Yinyan Wang, and Xingxing Wei.
\newblock Translation, scale and rotation: Cross-modal alignment meets
  rgb-infrared vehicle detection.
\newblock In {\em European Conference on Computer Vision}, pages 509--525.
  Springer, 2022.

\bibitem{napat2021misalignment}
Napat Wanchaitanawong, Masayuki Tanaka, Takashi Shibata, and Masatoshi Okutomi.
\newblock Multi-modal pedestrian detection with large misalignment based on
  modal-wise regression and multi-modal {IoU}.
\newblock In {\em International Conference on Machine Vision and Applications},
  pages 1--6, 2021.

\bibitem{tarvainen2017mean}
Antti Tarvainen and Harri Valpola.
\newblock Mean teachers are better role models: Weight-averaged consistency
  targets improve semi-supervised deep learning results.
\newblock {\em Advances in Neural Information Processing Systems}, 30, 2017.

\bibitem{kennerley20232pcnet}
Mikhail Kennerley, Jian-Gang Wang, Bharadwaj Veeravalli, and Robby~T Tan.
\newblock 2pcnet: Two-phase consistency training for day-to-night unsupervised
  domain adaptive object detection.
\newblock In {\em Proceedings of the IEEE Conference on Computer Vision and
  Pattern Recognition}, pages 11484--11493, 2023.

\bibitem{liu2022robust}
Chengxin Liu, Kewei Wang, Hao Lu, Zhiguo Cao, and Ziming Zhang.
\newblock Robust object detection with inaccurate bounding boxes.
\newblock In {\em European Conference on Computer Vision}, pages 53--69.
  Springer, 2022.

\bibitem{wu2023spatial}
Di~Wu, Pengfei Chen, Xuehui Yu, Guorong Li, Zhenjun Han, and Jianbin Jiao.
\newblock Spatial self-distillation for object detection with inaccurate
  bounding boxes.
\newblock In {\em Proceedings of the IEEE/CVF International Conference on
  Computer Vision}, pages 6855--6865, 2023.

\bibitem{zhou2020improving}
Kailai Zhou, Linsen Chen, and Xun Cao.
\newblock Improving multispectral pedestrian detection by addressing modality
  imbalance problems.
\newblock In {\em European Conference on Computer Vision}, pages 787--803.
  Springer, 2020.

\bibitem{liu2021swin}
Ze~Liu, Yutong Lin, Yue Cao, Han Hu, Yixuan Wei, Zheng Zhang, Stephen Lin, and
  Baining Guo.
\newblock Swin transformer: Hierarchical vision transformer using shifted
  windows.
\newblock In {\em Proceedings of the IEEE/CVF International Conference on
  Computer Vision}, pages 10012--10022, 2021.

\bibitem{sun2022drone}
Yiming Sun, Bing Cao, Pengfei Zhu, and Qinghua Hu.
\newblock Drone-based {RGB}-infrared cross-modality vehicle detection via
  uncertainty-aware learning.
\newblock {\em IEEE Transactions on Circuit System and Video Technology},
  32(10):6700--6713, 2022.

\bibitem{iafrcnn}
Chengyang Li, Dan Song, Ruofeng Tong, and Min Tang.
\newblock Illumination-aware faster {R-CNN} for robust multispectral pedestrian
  detection.
\newblock {\em Pattern Recognition}, 85:161--171, 2019.

\bibitem{guan2019fusion}
Dayan Guan, Yanpeng Cao, Jiangxin Yang, Yanlong Cao, and Michael~Ying Yang.
\newblock Fusion of multispectral data through illumination-aware deep neural
  networks for pedestrian detection.
\newblock {\em Information Fusion}, 50:148--157, 2019.

\bibitem{zhang2023tinet}
Yan Zhang, Huai Yu, Yujie He, Xinya Wang, and Wen Yang.
\newblock Illumination-guided rgbt object detection with inter-and
  intra-modality fusion.
\newblock {\em IEEE Transactions on Instrument and Measurement}, 72:1--13,
  2023.

\bibitem{kim2021uncertainty}
Jung~Uk Kim, Sungjune Park, and Yong~Man Ro.
\newblock Uncertainty-guided cross-modal learning for robust multispectral
  pedestrian detection.
\newblock {\em IEEE Transactions on Circuit System and Video Technology},
  32(3):1510--1523, 2021.

\bibitem{li2023multiscale}
Ruimin Li, Jiajun Xiang, Feixiang Sun, Ye~Yuan, Longwu Yuan, and Shuiping Gou.
\newblock Multiscale cross-modal homogeneity enhancement and confidence-aware
  fusion for multispectral pedestrian detection.
\newblock {\em IEEE Transactions on Multimedia}, 2023.

\bibitem{zhang2019cross}
Lu~Zhang, Zhiyong Liu, Shifeng Zhang, Xu~Yang, Hong Qiao, Kaizhu Huang, and
  Amir Hussain.
\newblock Cross-modality interactive attention network for multispectral
  pedestrian detection.
\newblock {\em Information Fusion}, 50:20--29, 2019.

\bibitem{shen2024icafusion}
Jifeng Shen, Yifei Chen, Yue Liu, Xin Zuo, Heng Fan, and Wankou Yang.
\newblock Icafusion: Iterative cross-attention guided feature fusion for
  multispectral object detection.
\newblock {\em Pattern Recognition}, 145:109913, 2024.

\bibitem{yuan2024c}
Maoxun Yuan and Xingxing Wei.
\newblock C 2 former: Calibrated and complementary transformer for rgb-infrared
  object detection.
\newblock {\em IEEE Transactions on Geoscience and Remote Sensing}, 2024.

\bibitem{yuan2024improving}
Maoxun Yuan, Xiaorong Shi, Nan Wang, Yinyan Wang, and Xingxing Wei.
\newblock Improving rgb-infrared object detection with cascade alignment-guided
  transformer.
\newblock {\em Information Fusion}, 105:102246, 2024.

\bibitem{chen2024weakly}
Chen Chen, Jiahao Qi, Xingyue Liu, Kangcheng Bin, Ruigang Fu, Xikun Hu, and
  Ping Zhong.
\newblock Weakly misalignment-free adaptive feature alignment for uavs-based
  multimodal object detection.
\newblock In {\em Proceedings of the IEEE Conference on Computer Vision and
  Pattern Recognition}, pages 26836--26845, 2024.

\bibitem{yi2019probabilistic}
Kun Yi and Jianxin Wu.
\newblock Probabilistic end-to-end noise correction for learning with noisy
  labels.
\newblock In {\em Proceedings of the IEEE Conference on Computer Vision and
  Pattern Recognition}, pages 7017--7025, 2019.

\bibitem{zheng2021meta}
Guoqing Zheng, Ahmed~Hassan Awadallah, and Susan Dumais.
\newblock Meta label correction for noisy label learning.
\newblock In {\em Proceedings of the AAAI conference on artificial
  intelligence}, volume~35, pages 11053--11061, 2021.

\bibitem{chadwick2019training}
Simon Chadwick and Paul Newman.
\newblock Training object detectors with noisy data.
\newblock In {\em IEEE Intelligent Vehicles Symposium (IV)}, pages 1319--1325.
  IEEE, 2019.

\bibitem{li2020learning}
Hengduo Li, Zuxuan Wu, Chen Zhu, Caiming Xiong, Richard Socher, and Larry~S
  Davis.
\newblock Learning from noisy anchors for one-stage object detection.
\newblock In {\em Proceedings of the IEEE Conference on Computer Vision and
  Pattern Recognition}, pages 10588--10597, 2020.

\bibitem{li2022cross}
Yu-Jhe Li, Xiaoliang Dai, Chih-Yao Ma, Yen-Cheng Liu, Kan Chen, Bichen Wu,
  Zijian He, Kris Kitani, and Peter Vajda.
\newblock Cross-domain adaptive teacher for object detection.
\newblock In {\em Proceedings of the IEEE Conference on Computer Vision and
  Pattern Recognition}, pages 7581--7590, 2022.

\bibitem{he2022cross}
Mengzhe He, Yali Wang, Jiaxi Wu, Yiru Wang, Hanqing Li, Bo~Li, Weihao Gan, Wei
  Wu, and Yu~Qiao.
\newblock Cross domain object detection by target-perceived dual branch
  distillation.
\newblock In {\em Proceedings of the IEEE Conference on Computer Vision and
  Pattern Recognition}, pages 9570--9580, 2022.

\bibitem{chen2023confidence}
Yajie Chen, Xin Yang, and Xiang Bai.
\newblock Confidence-weighted mutual supervision on dual networks for
  unsupervised cross-modality image segmentation.
\newblock {\em Science China Information Sciences}, 66(11):210104, 2023.

\bibitem{zhai2024maximizing}
Yiming Zhai, Chuanxian Ren, Youwei Luo, and Daoqing Dai.
\newblock Maximizing conditional independence for unsupervised domain
  adaptation.
\newblock {\em Science China Information Sciences}, 67(5):152108, 2024.

\bibitem{peng2022balanced}
Xiaokang Peng, Yake Wei, Andong Deng, Dong Wang, and Di~Hu.
\newblock Balanced multimodal learning via on-the-fly gradient modulation.
\newblock In {\em Proceedings of the IEEE Conference on Computer Vision and
  Pattern Recognition}, pages 8238--8247, 2022.

\bibitem{rezatofighi2019generalized}
Hamid Rezatofighi, Nathan Tsoi, JunYoung Gwak, Amir Sadeghian, Ian Reid, and
  Silvio Savarese.
\newblock Generalized intersection over union: A metric and a loss for bounding
  box regression.
\newblock In {\em Proceedings of the IEEE Conference on Computer Vision and
  Pattern Recognition}, pages 658--666, 2019.

\bibitem{mmdetection}
Kai Chen, Jiaqi Wang, Jiangmiao Pang, Yuhang Cao, and Yu~Xiong.
\newblock {MMDetection}: Open mmlab detection toolbox and benchmark.
\newblock {\em arXiv preprint arXiv:1906.07155}, 2019.

\bibitem{zhou2022mmrotate}
Yue Zhou, Xue Yang, Gefan Zhang, Jiabao Wang, Yanyi Liu, Liping Hou, Xue Jiang,
  Xingzhao Liu, Junchi Yan, Chengqi Lyu, Wenwei Zhang, and Kai Chen.
\newblock Mmrotate: A rotated object detection benchmark using pytorch.
\newblock In {\em ACM International Conference on Multimedia}, 2022.

\bibitem{pytorch}
Adam Paszke, Sam Gross, Francisco Massa, Adam Lerer, James Bradbury, Gregory
  Chanan, Trevor Killeen, Zeming Lin, Natalia Gimelshein, Luca Antiga, and
  Alban Desmaison.
\newblock Pytorch: An imperative style, high-performance deep learning library.
\newblock In {\em Advances in Neural Information Processing Systems}, pages
  8024--8035. 2019.

\bibitem{zhang2020atss}
Shifeng Zhang, Cheng Chi, Yongqiang Yao, Zhen Lei, and Stan~Z Li.
\newblock Bridging the gap between anchor-based and anchor-free detection via
  adaptive training sample selection.
\newblock In {\em Proceedings of the IEEE Conference on Computer Vision and
  Pattern Recognition}, pages 9759--9768, 2020.

\bibitem{broedermann2023hrfuser}
Tim Broedermann, Christos Sakaridis, Dengxin Dai, and Luc Van~Gool.
\newblock Hrfuser: A multi-resolution sensor fusion architecture for 2d object
  detection.
\newblock In {\em IEEE International Conference on Intelligent Transportation
  Systems}, pages 4159--4166. IEEE, 2023.

\bibitem{liu2016multispectral}
Jingjing Liu, Shaoting Zhang, Shu Wang, and Dimitris~N Metaxas.
\newblock Multispectral deep neural networks for pedestrian detection.
\newblock In {\em European Conference on Computer Vision}, pages 1--13, 2016.

\bibitem{he2023multispectral}
Xiao He, Chang Tang, Xin Zou, and Wei Zhang.
\newblock Multispectral object detection via cross-modal conflict-aware
  learning.
\newblock In {\em ACM International Conference on Multimedia}, pages
  1465--1474, 2023.

\end{thebibliography}

\end{document}